\documentclass{article}

\usepackage[final,nonatbib]{neurips_2023}

\usepackage[utf8]{inputenc} 
\usepackage[T1]{fontenc}    
\usepackage{hyperref}       
\usepackage{url}            
\usepackage{booktabs}       
\usepackage{amsfonts}       
\usepackage{nicefrac}       
\usepackage{microtype}      
\usepackage{xcolor}         

\title{Predictive Minds: LLMs As Atypical Active Inference Agents}

\author{
  Jan Kulveit\textsuperscript{1}\thanks{ \texttt{jk@acsresearch.org}} \quad\quad
  Clem von Stengel\textsuperscript{1} \quad\quad Roman Leventov\textsuperscript{2} \\
  \\
  \textsuperscript{1}
  Alignment of Complex Systems Research Group, Center for Theoretical Study, Charles University \\ 
   \textsuperscript{2} Gaia Consortium
}

\begin{document}

\maketitle

\begin{abstract}
  Large language models (LLMs) like GPT are often conceptualized as passive predictors, simulators, or even 'stochastic parrots'. We instead conceptualize LLMs by drawing on the theory of active inference originating in cognitive science and neuroscience. We examine similarities and differences between traditional active inference systems and LLMs, leading to the conclusion that, currently, LLMs lack a tight feedback loop between acting in the world and perceiving the impacts of their actions, but otherwise fit in the active inference paradigm. We list reasons why this loop may soon be closed, and possible consequences of this including enhanced model self-awareness and the drive to minimize prediction error by changing the world. 
\end{abstract}

\section{Introduction}

Foundation models, particularly large language Models (LLMs) like GPT \cite{brown2020language}, stand out as the most advanced general AI systems to date \cite{bubeck2023sparks}. LLMs are often perceived as mere predictors, primarily due to their training objective minimizing their loss on next-token prediction \cite{parrots}. This objective has led to the assumption that these models are inherently passive: designed to await prompts and respond without any real understanding of the world or implicit intention to influence or interact with the world. 
The theory of active inference, originating in cognitive science and neuroscience, offers an alternative viewpoint \cite{aifbook}. Active inference posits that biological systems like the human brain constantly update their internal models based on interactions with the environment, striving to minimize the difference between predicted and actual sensory inputs (a process also known as predictive processing) \cite{aifbook}. A fundamental tenet of active inference is that, in biological systems, this same objective also governs action: the system minimizes the difference between predicted and actual sensory input by actively altering its environment. 

This paper explores the intriguing possibility that LLMs, while predominantly seen as passive entities, might converge upon active inference agents closer to biological ones. We explore the parallels and distinctions between generative models like LLMs and those studied in active inference, and shed light on the emergent control loops that might arise, the incentives driving these changes, and the significant societal ramifications of such a shift.

\section{Background and related work}

\subsection{Conceptualizing LLMs}
\label{concept}

There have been various attempts to conceptualize LLMs, explain "how they actually work", and understand them using existing frameworks from a variety of fields.

One class of conceptualization focuses on the fact that the LM training objective is to minimize predictive loss, and the fact LLMs are not embodied in a way comparable to humans, but trained on large datasets of text from the internet. Bender et al. coined the term 'stochastic parrots' and claim that text generated by an LM is not grounded in communicative intent, any model of the world, or any model of the reader's state of mind \cite{parrots}. In a similar spirit, using framing from linguistics, Mahowald et al. conceptualize LLMs as models that are good at formal linguistic competence but incomplete at functional linguistic competence. According to this view, LLMs are good models of language but incomplete models of human thought, good at generating coherent, grammatical, and seemingly meaningful paragraphs of text, but failing in functional competence, which recruits multiple extralinguistic capacities that comprise human thought, such as formal reasoning, world knowledge, situation modeling, and social cognition \cite{mahowald2023}.

These reductionist views of LLMs were subject to considerable criticism. Mitchell and Krakauer, surveying the debate, note an opposing faction which argues that these networks truly understand language, can perform reasoning in a general way, and in a real sense understand concepts and capture important aspects of meaning \cite{mitchell}.  Mitchell and Krakauer's overall conclusion is that cognitive science is currently inadequate for answering such questions about LLMs.

Other conceptualizations of LLMs recognize that the trained model is a distinct object from the training process, and so that the nature of the training objective need not be shared by the resulting artifact. For example, based on experiments with LLMs autoregressively completing complex token sequences, Mirchandani et al. look at LLMs as general pattern machines, or general sequence modellers, driven by in-context learning \cite{mirchandani2023}. Others extend the 'general sequence modeling' in the direction of 'general computation'. For example, Guo et al. propose using natural language as a new programming language to describe task procedures, making them easily understandable to both humans and LLMs; they note that LLMs are capable of directly generating and executing natural language programs. In this conceptualization, trained LLMs are natural-language computers \cite{guo2023}.

Another conceptualization of LLMs, originating in the AI alignment community, views LLMs as general \textit{simulators} - simulating a learned distribution with various degrees of fidelity, which in the case of language models trained on a large corpus of text, is the mechanics underlying the genesis of the text, and so indirectly the world \cite{simulators}. This view explicitly assumes that LLMs learn world models, abstractions, algorithms to better model sequences. Similarly, Hubinger et. al. discusses how to understand LLMs as predictive models, and potential risks from such systems \cite{hubinger2023}. 

While not directly aimed at explaining how LLMs work, Lee et al. provide important context for this work, focusing on evaluating LLMs in interactive settings, and criticizing the fact that almost all benchmarks impose the non-interactive view, of models as passive predictors\cite{lee2022evaluating}.

\subsection{Active inference and predictive processing}

Originating in cognitive science and neuroscience, active inference offers a fresh lens through which to view cognitive processes. At its core, the theory suggests that living systems, such as animals or human brains, are in a constant state of updating their internal models \textit{while} acting on the environment, and both processes should be understood as minimizing the difference between predicted and actual sensory inputs (or, alternatively, variational free energy) \cite{aifbook}.

As an all-encompassing framework for building theories of cognitive systems, active inference should be compatible not only with process theories of brain function based on neurons \cite{friston2017active}, but also with a range of other computational structures (used to represent the world model), and a range of optimization procedures (used to minimize the difference between predicted and actual sensory inputs).  This makes active inference applicable - at least in principle - not only to humans and animals, but to a very broad range of systems, including the artificial. 

This naturally leads to our attempt to understand LLMs using the active inference framework. Pezzulo et al. compare active inference systems and "generative AIs" and claim that while both generative AI and active inference are based on generative models, they acquire and use them in fundamentally different ways.  Living organisms and active inference agents learn their generative models by engaging in purposive interactions with the environment and by predicting these interactions. The key difference is that learning and meaning is
grounded in sensorimotor experience,  providing biological agents with a core understanding and a sense of mattering upon which their subsequent knowledge and decisions are grounded \cite{pezzulo}. In the present work, we argue that this distinction is not necessarily as fundamental as assumed by Pezzulo et al., and may mostly disappear in the near future with tighter feedback loop between actions and observations.

\section{Similarities and differences between active inference systems and LLMs}

If we look at LLMs in the simulators framework and the active inference framework, we can note a number of similarities -- or even cases where the AI community and the active inference community describe the same phenomena using different terminology. In both cases, systems are described as equipped with a generative model able to simulate the system's sensory inputs. This model is updated in such a way that minimises prediction error - the difference between observed and simulated inputs. This process has been shown to be a form of approximate Bayesian inference in both the active inference \cite{aifbook, friston2013actinf} and LLM \cite{mingard2020sgd, xie2021explanation} literatures. 

\subsection{Predictions based on conceptualizing LLMs as special case of active inference systems}

The active inference conceptualization leads to a number of predictions, some of which are possible to verify experimentally using interoperability techniques.

Possibly the most striking one is obvious in hindsight: active inference postulates that the simple objective of minimizing prediction error is sufficient for learning complex world representations, behaviours and abstraction power, given a learning system with sufficient representation capacity. In predictive processing terminology, we can make an analogy between "perception" and the training process of LLMs: LLMs are fed texts from the internet and build generative models of the input. Because language is a reflection of the world, these models necessarily implicitly model not only language, but also the broader world. Therefore, we should expect LLMs to also learn complex world representations, abstractions, and the ability to simulate other systems, (given sufficient representation capacity). This is in contrast to the conceptualizations referenced in section \ref{concept}, which often predict that systems trained to predict next input are fundamentally limited, never able to generalize, unable to comprehend meaning, etc. Recent research has provided substantial evidence supporting the more optimistic view that large language models (LLMs) are analogous to biological systems at least in their ability to develop an emergent world model \cite{li2022emergent}, rich abstractions and the ability to predict general sequences \cite{mirchandani2023}. 

Another topic easier to understand through an active inference lens are hallucinations: where LLMs produce false or misleading information and present it as fact \cite{manakul2023gpthallucination}. Active inference claims that human perception is itself 'constrained hallucination'\cite{parr2021understanding}, where our predictions about sensory inputs are constantly synchronized with reality through the error signal, propagated backwards. 
In this perspective, the data on which LLMs are trained could be understood as sensory input. What's striking about these inputs is, in contrast to human sensory inputs, the data are \textit{not} based on perceiving reality from one specific perspective in one point of time. Quite the opposite: for an intuitive understanding of the nature of the data LLMs are trained on, imagine that your own sensory input was exhausted by overhearing human conversations, with the caveat that what you hear every few minutes randomly switches between conversations taking place out of order in different years, contexts and speakers. In contrast to the typical human situation - trying to predict what you would hear next - you would often need to entertain \textit{many} different hypotheses about the current context. For example, consider hearing someone say "And she drew her sword and exclaimed 'Heretics must die!'". When attempting to predict the continuation, it seems necessary to entertain many possibilities - such as the context being a realistic description of some medieval world, or a fantasy tale, or someone playing a video-game. If a biological, brain-based active inference system was tasked with predicting such contextless words, then various fantasy and counterfactual worlds would seem as real as actual current affairs. In this conceptualization, some hallucinations in LLMs are not some sort of surprising failure mode of AI systems, but what you should expect from a system tasked to predict text with minimal context, not anchored to some specific temporal or contextual vantage point. Another striking feature of LLMs in deployment is that outputs of the generative model are not distinguished from inputs: the model's output becomes part of its own 'sensory' state. Intuitively, this would be similar to a human unable to distinguish between their own actions and external influences - which actually sometimes manifests as the psychiatric condition known as 'delusion of control' \cite{fletcher2009perceiving}.   

This frame suggests directions to make LLMs less prone to hallucinations: make the learning context of the LLM more situated and contextually stable (that is, present training documents in a more systematic fashion). Additionally, it could help to distinguish between completions by the model and inputs from the user, similar to the approach of Ortega et al. \cite{ortega2021shaking}.

\subsection{What is an LLM's actuator?}
\label{actuator}

One suggested fundamental difference between LLMs and active inference systems is the inherent passivity of LLMs - their inability to \textit{act} in the world \cite{pezzulo}. 
We argue that this is mostly a matter of degree and not a categorical difference. While LLMs don't have actuators in the physical world like humans or robots, they still have the ability to act, in the sense that their predictions do affect the world. In active inference terminology, LLM outputs could be understood as the 'action states' in the Markov blanket. These states have some effect on the world via multiple causal pathways, and the resulting changes can in principle influence its 'sensory states' - that is, various pieces of text on the internet and included in the training set. Some clear pathways:
\begin{enumerate}
    \item Direct inclusion of text generated by LLM in web pages.
    \item Human users asking LLM based assistants for plans and executing those plans in the world.
    \item Text input for a huge range of other software systems (LLMs as glue code and so-called "robotic process automation").
    \item Indirect influence on how humans think about things, e.g. learning about a concept from an LLM based assistance.
\end{enumerate}

Some of these effects are already studied in the ML literature, but mostly in the context of feedback loops amplifying bias \cite{taori2023data} or as an example of performative prediction \cite{perdomo2020performative}. Here, we propose a broader interpretation: understanding these effects as actions in the sense it takes in active inference. The nature of the medium through which LLMs "perceive" and "act" on the world, which is mostly text, should not obscure the fundamental similarity to active inference agents. We agree with McGregor's argument \cite{mcgregor2023chatgpt} that we should explicitly distinguish between two notions of embodiment: on the one hand, whether a system’s body is tangible or not, and on the other
hand, whether a system is physically situated or not (i.e. whether or not it interacts physically with
any part of the universe). LLMs are embodied in this second sense. In this view, interactions of LLMs with users in deployment are essentially 'actions'. Every token generated in conversation with users is a micro-action, and the sum of all of these actions do influence the world, and some of these changes get reflected in the input world (public texts on the internet). So, at least in principle, LLMs have one open causal path to bring the world of words closer to their predictions.

\subsection{Closing the action loop of active inference}

Given that the "not acting on the world" assumption of "LLMs as passive simulators" does not hold, the main current difference between LLMs and active inference systems is that LLMs mostly are not yet able to "perceive" the impacts of their actions. In other words, the loop between actions, external world states, and perceptions is not closed (or anyway is not fast). While living organisms constantly run both perception and action loops,  training new generations of an LLM happens only once a year or so - and the impacts of actions of the LLM currently mostly do not feed back into the new base model's training.

What would need to be changed for LLMs to perceive the results of their own actions, and thus close the “gap” between action and perception? The key piece is that the actions taken by an LLM after deployment, in the sense discussed in section \ref{actuator}, feed back into the training process of a future LLM. Furthermore, it is required that successive LLMs are sufficiently similar, and have sufficient representational capacity, such that they can “self-identify” with successive training iterations (see \cite{leventov} for a discussion of “the GPT lineage” as an agent). 


A minimal version of this can occur with in-context learning \cite{dai2022incontext}, real-time access to web search (as with Bing Chat and Google Bard), or a training environment in which the model can take actions which influence its reward (such as with GATO \cite{reed2022generalist}, or RLHF \cite{ouyang2022training}). However in each of these cases, there is no feedback from the actions taken during deployment and subsequent training of the LLM. There are three ways we foresee this happening in the near future:

\begin{enumerate}
    \item The outputs of a model are used to train a next generation model, e.g. through model outputs being published on the internet and not filtered out during data curation.
    \item The data collected from interactions with the models, such as from user conversations with a chatbot, are used in fine-tuning future versions of the same model.
    \item Continuous online learning, in which the outputs of a model and user responses are directly used as a training signal to update the model.
\end{enumerate}

Where these routes are in order of increasingly tight feedback loops (where "tighter" means on a shorter timescale, with consecutive generations sharing more of the earlier model's weights, and with the interaction forming a larger percentage of training data - increased bandwidth). 

We expect that there will be active effort by developers to close the feedback gap and make the action loop more prominent because of commercial incentives to make LLMs better at quickly adapting to new information, acting independently, or otherwise agent-like. Active inference as a theory of agency predicts closing the loop would naturally cause LLMs to become more agentic, emergently learning to change the world to more closely match the internal states (and thus predictions) of LLMs.

\section{Implications of active LLMs}

The evolution of LLMs into active agents would carry profound societal implications and risks. Using active inference as a theoretical framework to make predictions about such Active LLMs is a fruitful direction. We focus on emergence of increased self-awareness. 

\subsection{Enhancing model self-awareness}

A straightforward prediction of the active inference frame in this paper is that the described tightening of the feedback loop is likely to to augment and increase models' self-awareness. A recent study of self-awareness \cite{berglund2023taken} in LLMs emphasizes the importance of self-awareness from a safety perspective, but this work is overall uncertain about what stage of LLM training will be more important for the emergence of situational awareness in future models, and focuses on evaluating sophisticated out-of-context reasoning as a proxy of self-awareness. In contrast, the active inference literature emphasizes the importance of observing the consequences of one's own actions for developing functional self-awareness \cite[p.~112]{friston2019fep}.

As these loops tighten, we expect models to enhance in self-awareness by acquiring more information about themselves and observing the repercussions of their actions in the environment. Consider the self-localization problem discussed by \cite{berglund2023taken}. Construct a thought experiment in which a human faces a similar self-localization problem: assume, instead of one's usual sensory inputs, that the human is hooked to a stream of dozens of security cameras. To increase the human's ability to self-localize is to equip them with more information about their own appearance, for example, hair colour. A different, highly effective way to self-localize is via performing an action, for example by waving a hand.





\section{Conclusions}

By examining the learning objectives and feedback loops of active inference, in comparison to those of LLMs, we posited that LLMs can be understood as an unusual example of active inference agents with a gap in their feedback loop from action to perception. In this framework, their transition to acting in the world as living organisms do depends on their closing the gap between interacting (with users) and training.

The potential metamorphosis of LLMs into active LLMs could lead to more adaptive and self-aware AI systems, bearing substantial societal implications. The densification and acceleration of feedback loops could augment not only models' self-awareness but also lead to a drive to modify the world - driven purely by the prediction error minimization objective, without intentional effort to make the models more agent-like.

\section{Acknowledgements}

We thank Rose Hadshar and Gavin Leech for help with writing and editing, and Tomáš Gavenčiak, Simon McGregor and Nicholas Kees Dupuis for valuable discussions. JK and CvS were supported by PRIMUS grant from Charles University. GPT4 was used for editing the draft, simulating readers, and title suggestions.

\bibliography{actinf} 
\bibliographystyle{plain} 

\medskip

{
\small


\end{document}